\documentclass[11pt]{article}
\usepackage{coling2016}
\usepackage{times}
\usepackage{url}
\usepackage{latexsym}
\usepackage{graphicx}
\usepackage{enumitem}
\setitemize{noitemsep,topsep=1pt,partopsep=0.1pt,parsep=1pt}
\setlist[enumerate]{topsep=1pt,noitemsep,partopsep=0.1pt,parsep=1pt}
\newcommand{\clt}{\begin{small} \begin{list}{}{\setlength{\parsep}{-0.05mm}
\setlength{\itemsep}{-0.05mm} \setlength{\leftmargin}{4mm}} \item}
\newcommand{\clr}{\end{list} \end{small}}
\newcommand{\com}[1]{\textbf{#1}}
\newcommand{\cl}{\item}

\newcommand{\while}{\com{while }}

\title{Contradiction Detection for Rumorous Claims}

\author{    
 Piroska Lendvai\\
        Computational Linguistics\\
        Saarland University\\
        Saarbr\"ucken, Germany\\
        {\tt piroska.r@gmail.com}\\
      \And   
         Uwe D. Reichel\\
       Research Institute for Linguistics\\
        Hungarian Academy of Sciences\\
        Budapest, Hungary\\
        {\tt uwe.reichel@nytud.mta.hu}
  }

\date{}

\begin{document}
\maketitle
\begin{abstract} 
The utilization of social media material in journalistic workflows is increasing, demanding automated methods for the identification of mis- and disinformation. Since textual contradiction across social media posts can be a signal of rumorousness, we seek to model how claims in Twitter posts are being textually contradicted. We identify two different contexts in which contradiction emerges: its broader form can be observed across independently posted tweets and its more specific form in threaded conversations. We define how the two scenarios differ in terms of central elements of argumentation: claims and conversation structure. We  design and evaluate models for the two scenarios uniformly as 3-way Recognizing Textual Entailment tasks in order to  represent claims  and conversation structure implicitly in a generic inference  model, while previous studies used  explicit or no representation of these  properties. To address noisy text, our classifiers use simple  similarity features derived from the string and part-of-speech level. Corpus statistics reveal distribution differences for these features in contradictory as opposed to non-contradictory tweet relations, and the  classifiers yield state of the art performance.
\end{abstract}

\section{Introduction and Task Definition}
\label{sec:intro}
 
\begin{figure}[htbp]
\begin{center}
\includegraphics[scale=.32]{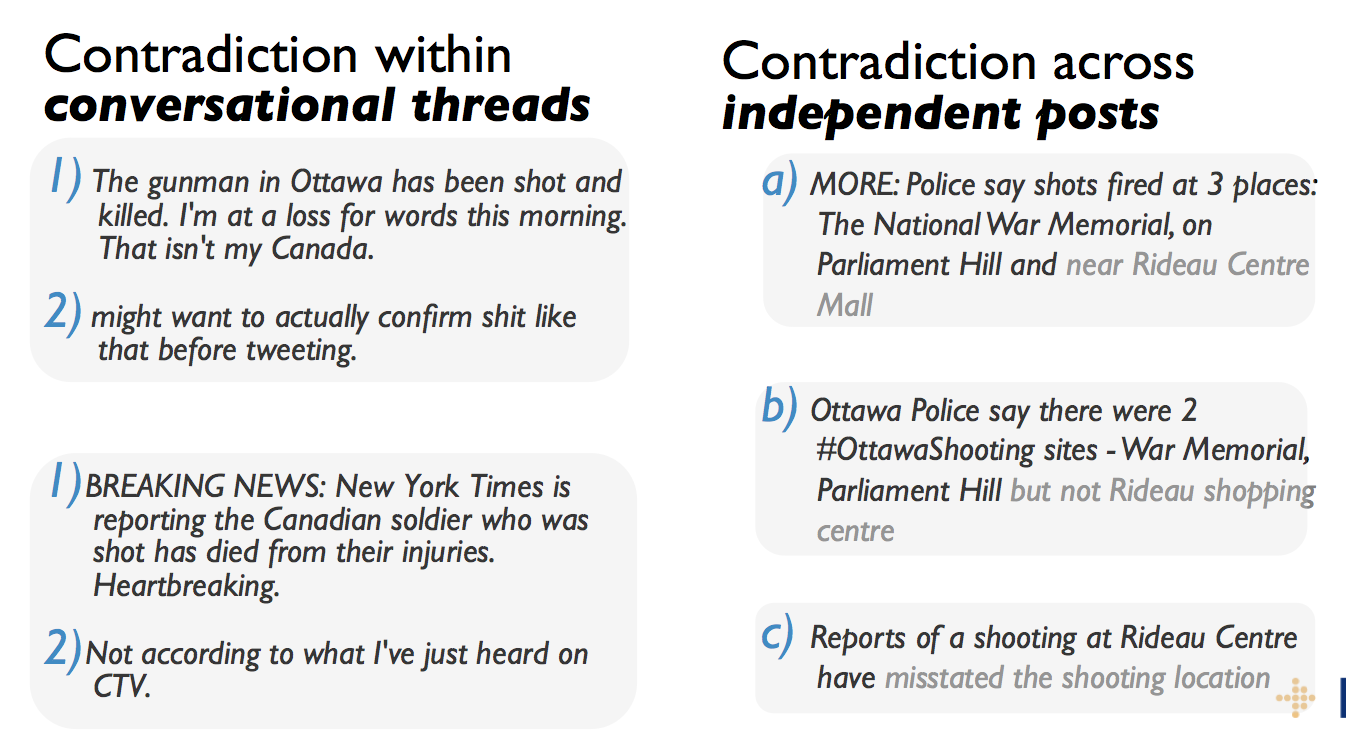}
\hfill
\includegraphics[scale=.32]{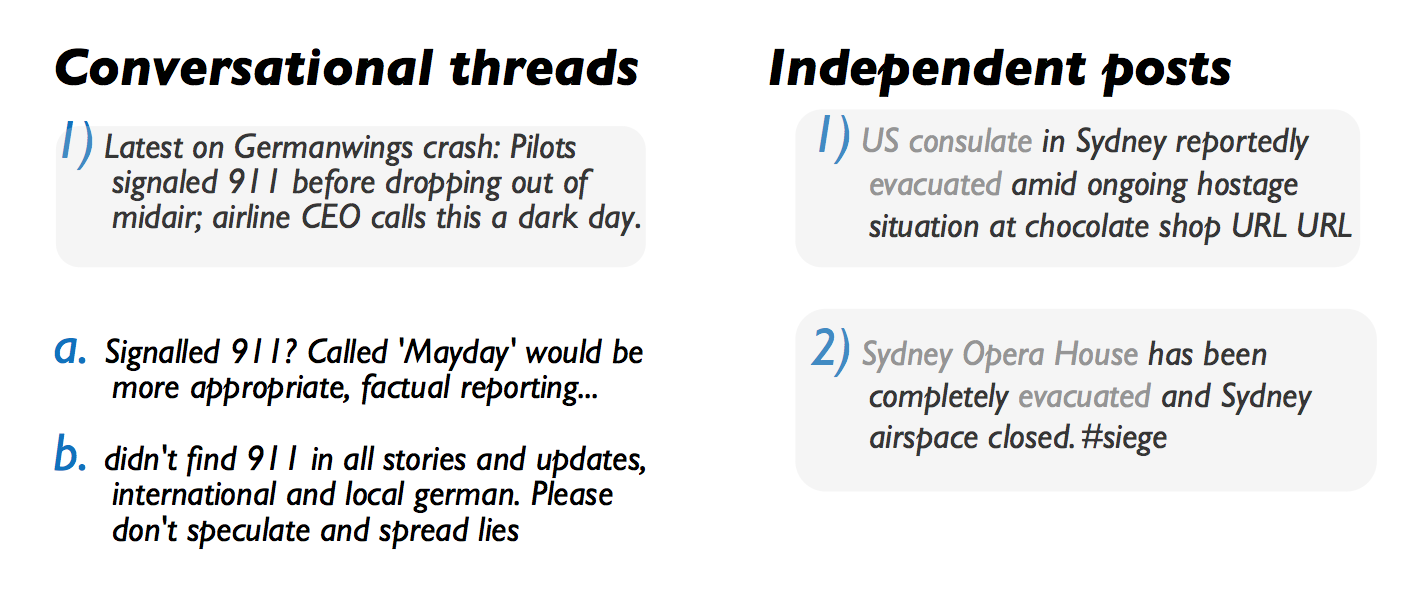}
\caption{Explicit (far left: in threads, left: in independent posts) vs implicit (right: in threads, far right: in independent posts) contradictions in threaded discussions and in independent posts.}
\label{fig:tweets}
\end{center}
\end{figure}

Assigning a veracity judgment to a claim appearing on social media requires complex procedures including reasoning on  claims aggregated from multiple microposts, to establish claim veracity status (resolved or not) and veracity value (true or false). Until resolution, a claim circulating on social media platforms is regarded as a rumor \cite{Mendoza2010}. The detection of  contradicting and disagreeing microposts   supplies important cues to claim veracity processing procedures. These tasks are  challenging to automatize not only due to the surface  noisiness and conciseness of user generated content. One  complicating factor is that claim denial or rejection is linguistically often not explicitly expressed, but appears without  classical rejection markers or  modality and speculation cues \cite{exprom}. Explicit and implicit contradictions furthermore arise in different contexts: in threaded discussions, but also across independently posted messages; both contexts are exemplified in Figure~\ref{fig:tweets} on Twitter data.

Language technology has not yet solved the processing of contradiction-powering phenomena, such as negation \cite{morante_scope} and stance detection \cite{StanceSemEval2016}, where stance is defined to express speaker favorability towards an evaluation target, usually an entity or concept.  In the veracity computation scenario we can speak of {\it claim targets} that are above the entity level:targets are entire rumors, such as '11 people died during the Charlie Hebdo attack'. Contradiction and stance detection have so far only marginally been addressed in the veracity context \cite{deMarneffe_didithappen,Emergent,Lukasik}.

We propose investigating the advantages of incorporating claim target and conversation context as premises in the Recognizing Textual Entailment (RTE) framework for contradiction detection in rumorous tweets. Our goals are manifold: (a) to offer richer context in contradiction modeling than what would be available on the level of individual tweets, the typical unit of analysis in previous studies; (b) to train and test supervised classifiers for contradiction detection in the RTE inference framework; (c) to address contradiction detection at the level of text similarity only, as opposed to semantic similarity \cite{2015semeval}; (d) to distinguish and focus on two different contradiction relationship types, each involving specific combinations of claim target mention,  polarity, and contextual proximity, in particular:

\begin{enumerate} 
\item {\bf Independent contradictions}: Contradictory relation between independent posts, in which two tweets  contain different information  about  the same claim target that cannot  simultaneously hold. The two messages are independently posted, i.e., not occurring within a structured conversation.
\item {\bf Disagreeing replies}: Contradictory relation between a claim-originating tweet and a direct reply to it, whereby the reply expresses disagreement with respect to the claim-introducing tweet.
\end{enumerate}

Contradiction between independently posted tweets typically arises in a broad discourse setting, and may feature larger distance in terms of  time, space, and source of information.
The claim target is mentioned in both posts in the contradiction pair, since these posts are  uninformed about each other or assume uninformedness of the reader, and thus do not or can not make coreference to their shared claim target.  
Due to the same reason, the  polarity of both posts with respect to the claim can be identical.
Texts paired in this type of contradiction resemble those of the recent Interpretable Semantic Similarity shared task  \cite{semeval_interpretable} that calls to identify five chunk level semantic relation types (equivalence, opposition, specificity, similarity  or  relatedness)
between two texts that originate from headlines or captions.
Disagreeing replies are more specific instances of contradiction:  contextual proximity is small and trivially identifiable 
by means of 
e.g. social media platform metadata, for example the property encoding the  tweet ID to which the reply was sent, which in our setup is always a thread-initiating tweet.  The claim target is by definition assumed to be contained in the thread-initiating tweet (sometimes termed as claim- or rumor-source tweet). It can be the case that the claim target is not contained in the reply, which can be explained by the proximity and thus shared context of the two posts.
The polarity values in source and reply must by definition be different; we refer to this scenario as Disagreeing replies.
Importantly, replies may not contain a (counter-)claim on their own 
but some other form to express disagreement and polarity -- for example in terms of speculative language use, 
or the presence of extra-linguistic cues such as a URL pointing to an online article that holds contradictory content.
Such cues are  difficult to decode for a machine, and their representation for training automatic classifiers is largely unexplored. 
Note that we do not make assumptions or restrictions about how the claim target is encoded textually in any of the two scenarios.

In this study, we tackle both contradiction types using a single generic approach: we recast them
as three-way 
RTE tasks on pairs of tweets. 
The findings of our previous study in which   semantic inference systems with sophisticated, corpus-based or manually created syntactico-semantic features were applied to contradiction-labeled data 
indicate the lack of robust syntactic and semantic analysis for short and noisy texts; cf. Chapter 3 in \cite{pheme_deliv_422}. This motivates our current simple text similarity metrics in search of  alternative  methods for the contradiction processing task.

In Section \ref{sec:PreviousWork} we introduce  related work and resources,  in Sections \ref{sec:data} and \ref{sec:features} present and motivate the collections and the features used for modeling.
After the description of  method and scores in Section \ref{sec:method},  findings are discussed in Section \ref{sec:conc}.

\section{Related work and resources}
\label{sec:PreviousWork}

\paragraph{Recognizing Textual Entailment (RTE)} Processing semantic inference phenomena  such as contradiction, entailment and stance between text pairs has been gaining momentum in language technology.  
Inference has been suggested to be conveniently formalized in the generic framework of RTE\footnote{http://www.aclweb.org/aclwiki/index.php?title=Recognizing\_Textual\_Entailment} \cite{dagan_pascal}. 
As an improvement over the binary Entailment vs Non-entailment scenario,
three-way RTE has appeared but is still scarcely investigated \cite{Emergent,Lendvaietal_pheme_rte}. 
The {\it Entailment} relation between two text snippets holds if the claim present in snippet B can be concluded from snippet A. The {\it Contradiction} relation applies when the claim in A and the claim in B cannot be simultaneously true. The {\it Unknown} relation applies if A and B neither entail nor contradict each other.

The RTE-3 benchmark dataset is the first resource that labels paired text snippets in terms of 3-way RTE judgments \cite{deMarneffe}, but it is comprised of general newswire texts.  
Similarly, the new large annotated  corpus used  for deep models for  entailment ~\cite{Bowman}
labeled text pairs as Contradiction are too broadly defined, i.e., expressing generic semantic incoherence  rather than semantically motivated polarization and mismatch that we are after, which questions its utility in the rumor verification context.

 As far as contradiction processing is concerned,  accounting for negation in RTE is the focus of a recent study \cite{madhumita}, but it is still set according to the binary RTE setup.
 A standalone contradiction detection system was implemented by  \cite{deMarneffe}, using complex rule-based features.
A specific RTE application, the Excitement Open Platform\footnote{http://hltfbk.github.io/Excitement-Open-Platform} \cite{pado_eop} has been developed to provide a generic platform for applied RTE. It integrates several entailment decision algorithms, while only the Maximum Entropy-based model \cite{wang2007recognizing} is available for 3-way RTE classification.  This model implements state-of-the-art linguistic preprocessing  augmented with  lexical resources (WordNet, VerbOcean), and uses the output of part-of-speech and dependency parsing in its structure-oriented, overlap-based approach for classification and was tested for both our tasks as explained in \cite{pheme_deliv_422}.

\paragraph{Stance detection}
Stance classification and stance-labeled corpora are relevant for contradiction detection, because  the relationship of two  texts expressing opposite stance (positive and negative) can in some contexts be judged to be contradictory: this is exactly what our Disagreeing reply scenario covers.
 Stance classification for rumors was introduced  by 
\cite{Qazvinian2011} where the goal was to generate a  binary (for or against) stance judgment. 
 Stance is typically classified on the level of individual tweets: reported  approaches predominantly utilize  statistical models, involving supervised machine learning \cite{deMarneffe_didithappen} and RTE \cite{Emergent}.
Another relevant aspect of stance  detection for our current study is the presence of the stance target in the text to be stance-labeled.
A recent shared task on social media data  defined separate challenges depending on whether
target-specific training data is included in the task or not  \cite{StanceSemEval2016}; the latter requires additional effort to encode  information about the stance target, cf. e.g. \cite{Augenstein2016SemEval}.  
The PHEME project released a new stance-labeled social media dataset \cite{Zubiaga} that we also utilize
as described next.

\section{Data}
\label{sec:data}

The two datasets corresponding to our two tasks are drawn from a freely available, annotated social media corpus\footnote{https://figshare.com/articles/PHEME\_rumour\_scheme\_dataset\_journalism\_use\_ case/2068650} that was collected from the Twitter platform\footnote{twitter.com} via filtering on event-related keywords and hashtags in the Twitter Streaming API.  We worked with English tweets related to four events: the Ottawa shooting\footnote{https://en.wikipedia.org/wiki/2014\_shootings\_at\_Parliament\_Hill,\_Ottawa}, the Sydney Siege\footnote{https://en.wikipedia.org/wiki/2014\_Sydney\_hostage\_crisis},  the Germanwings crash\footnote{https://en.wikipedia.org/wiki/Germanwings\_Flight\_9525}, and the Charlie Hebdo shooting\footnote{https://en.wikipedia.org/wiki/Charlie\_Hebdo\_shooting}.  
Each event in the corpus was pre-annotated as explained in  \cite{Zubiaga} for several rumorous claims\footnote{{\it Rumor}, {\it rumorous claim} and {\it claim} are used interchangeably throughout the paper to refer to the same concept.} -- officially not yet confirmed statements lexicalized by a concise proposition, e.g. 
{\sf  \small "Four cartoonists were killed in the Charlie Hebdo attack"} and
{\sf  \small "French media outlets to be placed under police protection"}.
The corpus collection method was based on a retweet threshold, therefore most tweets originate from authoritative sources using relatively well-formed language, whereas replying tweets 
often feature non-standard language use. 

\begin{table}[t]
\centering
\begin{small}
\begin{tabular}{|l |r r r r r|| r r r r r|}
\hline 
\bf \small event &\bf \small  ENT  & \bf \small CON & \bf \small UNK   &  \bf \small  \#uniq &  \bf \small  \#uniq & \bf \small  ENT  & \bf \small CON & \bf \small UNK   &  \bf \small  \#uniq &  \bf \small  \#uniq\\ 
   &   &   &  & \bf \small clms &  \bf \small tws  &   &   &  & \bf \small clms &  \bf \small tws\\ 
\hline
\hline
chebdo & 143&34&486  & 36&736  & 647 &  427 &  866 & 27 & 199   \\
gwings &39&6& 107& 13&176 & 461 &  257 &  447 & 4 & 29  \\
ottawa&79&37&292 & 28&465 &  555 & 377 & 168  & 18 & 125  \\
ssiege &112&59& 456& 37&697  &   332 &  317 &  565 & 21 & 143 \\
\hline
\hline
&373&136& 1341&114 &2074 &1995  &  1378 &  2046 & 70 &  496\\
\hline
\end{tabular}
\caption{{\it Threads} (left) and {\it iPosts} (right) RTE datasets compiled from 4 crisis events: amount of pairs per entailment type ({\it ENT, CON, UNK}), amount of unique rumorous claims ({\it \#uniq clms}) used for creating the pairs, amount of unique tweets discussing these claims ({\it \#uniq tws}).}
\label{tab:collections}
\end{small}
\end{table}

Tweets are organized into threaded conversations in the corpus and are marked up with respect to stance, certainty, evidentiality, and other veracity-related properties; for full details on released data we refer to \cite{Zubiaga}. The  dataset on which we run  disagreeing reply detection  (henceforth:  {\it Threads}) was converted by us to RTE format based on  the threaded conversations labeled in this corpus.
We created the Threads RTE dataset drawing on manually pre-assigned Response Type labels by  \cite{Zubiaga} that were meant to characterize source tweet -- replying tweet relations in terms of four categories. We mapped these four categories onto three RTE labels: a reply pre-labeled as {\it Agreed} with respect to its source tweet was mapped to {\it Entailment},  a reply pre-labeled as {\it Disagreed} was mapped to  {\it Contradiction}, while replies pre-labeled as {\it AppealforMoreInfo} and {\it Comment} were mapped to {\it Unknown}.
Only direct replies to source tweets relating to the same four events as in the independent posts RTE dataset were kept.
There are 1,850 tweet pairs in this set; the proportion of contradiction instances amounts to 7\%.
The  {\it Threads} dataset holds {\it CON}, {\it ENT} and {\it UNK} pairs 
as exemplified below.  Conform the RTE format,   pair elements are termed  {\it text} and  {\it hypothesis} -- note that directionality between  {\it t} and {\it h} is assumed as symmetric in our current context so  {\it t} and {\it h} are assigned based on token-level length.

\begin{itemize}
\item {\bf CON} $<$t$>${\sf  \small We understand there are two gunmen and up to a dozen hostages inside the cafe under siege at Sydney.. ISIS flags remain on display 7News}$<$/t$>$
$<$h$>${\sf  \small not ISIS flags}$<$/h$>$

\item  {\bf ENT} $<$t$>${\sf  \small Report: Co-Pilot Locked Out Of Cockpit Before Fatal Plane Crash URL Germanwings URL}$<$/t$>$
$<$h$>${\sf  \small This sounds like pilot suicide.}$<$/h$>$

\item   {\bf UNK} $<$t$>${\sf  \small BREAKING NEWS: At least 3 shots fired at Ottawa War Memorial. One soldier confirmed shot - URL URL}$<$/t$>$
$<$h$>${\sf  \small All our domestic military should be armed, now.}$<$/h$>$.
\end{itemize}

The independently posted tweets dataset (henceforth: {\it iPosts}) that we used for contradiction detection between independently emerging claim-initiating tweets is described in \cite{Lendvaietal_pheme_rte}. This collection is holds 5.4k RTE pairs generated from about 500 English tweets using semi-automatic 3-way RTE labeling, based on semantic or numeric mismatches between the rumorous claims annotated in the data. The proportion of contradictory pairs ({\it CON}) amounts to 25\%. 
The two collections are quantified in Table \ref{tab:collections}. {\it iPosts} dataset examples are given below.

\begin{itemize}
\item  {\bf CON} $<$t$>${\sf  \small 12 people now known to have died after gunmen stormed the Paris HQ of magazine CharlieHebdo URL URL}$<$/t$>$
$<$h$>${\sf  \small Awful. 11 shot dead in an assault on a Paris magazine. URL CharlieHebdo URL}$<$/h$>$

\item  {\bf ENT} $<$t$>${\sf  \small SYDNEY ATTACK - Hostages at Sydney cafe - Up to 20 hostages - Up to 2 gunmen - Hostages seen holding ISIS flag DEVELOPING..}$<$/t$>$
$<$h$>${\sf  \small Up to 20 held hostage in Sydney Lindt Cafe siege URL URL}$<$/h$>$

\item  {\bf UNK} $<$t$>${\sf  \small BREAKING: NSW police have confirmed the siege in Sydney's CBD is now over, a police officer is reportedly among the several injured.}$<$/t$>$
$<$h$>${\sf  \small Update: Airspace over Sydney has been shut down. Live coverage: URL sydneysiege}$<$/h$>$.
\end{itemize}

\section{Text similarity features}
\label{sec:features}

Data preprocessing on both  datasets included screen name and hashtag sign removal and  URL masking.  Then, for each tweet pair we extracted vocabulary overlap and local text alignment features. The tweets were part-of-speech-tagged using the Balloon toolkit \cite{ReichelIS2012} (PENN tagset, \cite{Penn1999}), normalized to lowercase  and stemmed using an adapted version of the Porter stemmer \cite{Porter1980}. Content words were defined to belong to the set of nouns, verbs, adjectives, adverbs, and numbers, and were identified by their part of speech labels. All punctuation was removed.

\subsection{Vocabulary overlap}

Vocabulary overlap was calculated for content word stem types in terms of the Cosine similarity and the F1 score.
The Cosine similarity of two tweets is defined as
$C(X,Y) = \frac{|X \cap Y|}{\sqrt{|X|\cdot|Y|}}$, where $X$ and $Y$ denote the sets of content word stems in the tweet pair.

The F1 score is defined as the harmonic mean of precision and recall. Precision and recall here refer to covering the vocabulary $X$ of one tweet by the vocabulary $Y$ of another tweet (or vice versa). It is given by 
$F1 = 2 \cdot \frac{\frac{|X \cap Y|}{|X|} \cdot \frac{|X \cap Y|}{|Y|}}{\frac{|X \cap Y|}{|X|} + \frac{|X \cap Y|}{|Y|}}$. Again the vocabularies $X$ and $Y$ consist of stemmed content words. Just like the Cosine index, the F1 score is a symmetric similarity metric.

These two metrics are additionally applied to the content word POS label inventories within the tweet pair, which gives the four features {\em cosine, cosine\_pos, f\_score}, and {\em f\_score\_pos}, respectively.

\subsection{Local alignment}

The amount of stemmed word token overlap was measured by applying local alignment of the token sequences using the Smith-Waterman algorithm \cite{SmithJMB1981}. We chose a score function rewarding zero substitutions by $+1$, and punishing insertions, deletions, and substitutions each by $0$-reset. Having filled in the score matrix $H$, alignment was iteratively applied the following way:

\clt \while $\max(H) \geq t$
\clt -- trace back from the cell containing this maximum the path
leading to it until a zero-cell is reached
\cl -- add the substring collected on this way to the set of aligned substrings
\cl -- set all traversed cells to 0.
\clr
\clr

The threshold
$t$ defines the required minimum length of aligned substrings. It is
set to 1 in this study, thus it supports a complete alignment of
any pair of permutations of $x$. The traversed cells are set to
$0$ after each iteration step to prevent that one substring would be
related to more than one alignment pair. This approach would allow for two restrictions: to prevent cross alignment not just the traversed cells $[i,j]$ but for each of these cells its entire row $i$ and column $j$ needs to be set to 0. Second, if only the longest common substring is of interest, then the iteration is trivially to be stopped after the first step. Since we did not make use of these restrictions, in our case the alignment supports cross-dependencies and can be regarded as an iterative application of a longest common substring match.

From the substring pairs in tweets $x$ and $y$ aligned this way, we extracted two text similarity measures:

\begin{itemize}
\item {\em laProp}: the proportion of locally aligned tokens over both tweets $\frac{m(x)+m(y)}{n(x)+n(y)}$
\item {\em laPropS}: the proportion of aligned tokens in the shorter tweet 
$\frac{m(\hat{z})}{n(\hat{z})}, \hat{z} = \arg\min_{z \in \{x,y\}} [n(z)]$,
\end{itemize}

where $n(z)$ denotes the number of all tokens and $m(z)$ the number of aligned tokens in tweet $z$.

\subsection{Corpus statistics}
\label{sec:corpstat}

\begin{figure}[!ht]
\begin{center}
\includegraphics[scale=.4]{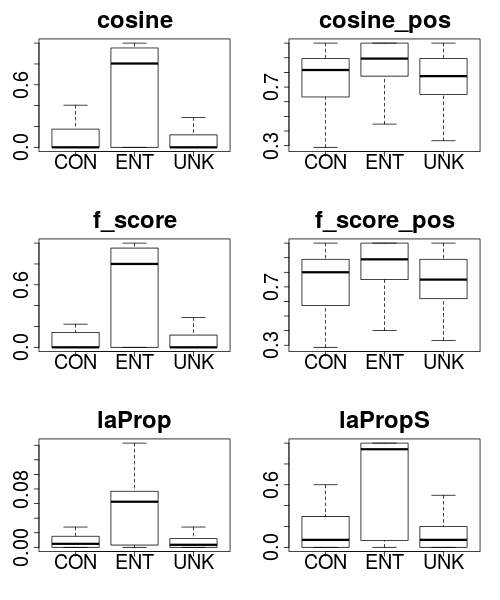}
\hfill
\includegraphics[scale=.4]{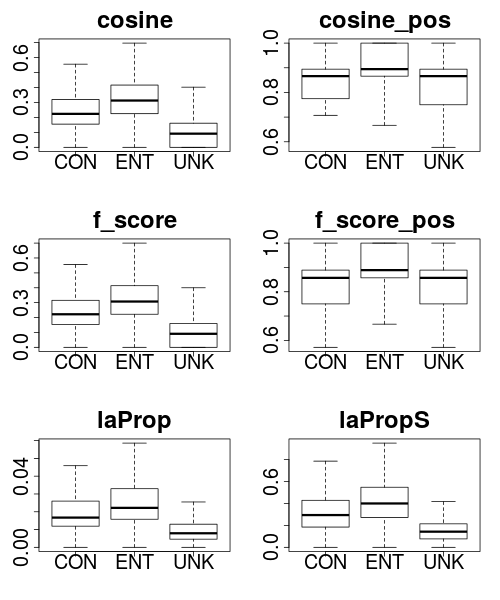}
\caption{Distributions of the similarity metrics by tweet pair class for the event {\em chebdo} in the {\em Threads} ({\bf left}) and the {\em iPosts} dataset (\bf right).}
\label{fig:stat1}
\end{center}
\end{figure}

\begin{figure}[!ht]
\begin{center}
\includegraphics[scale=.4]{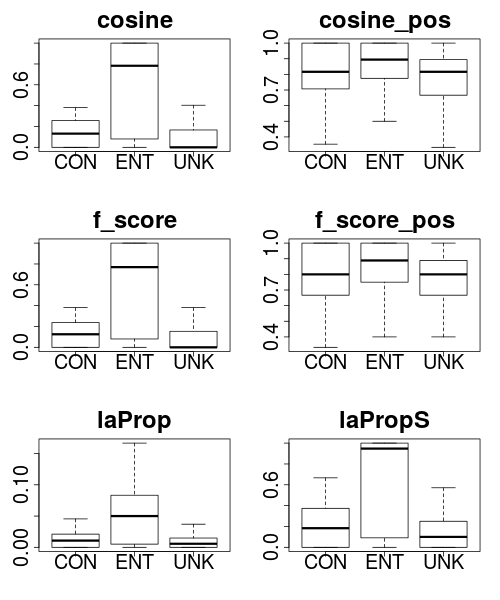}
\hfill
\includegraphics[scale=.4]{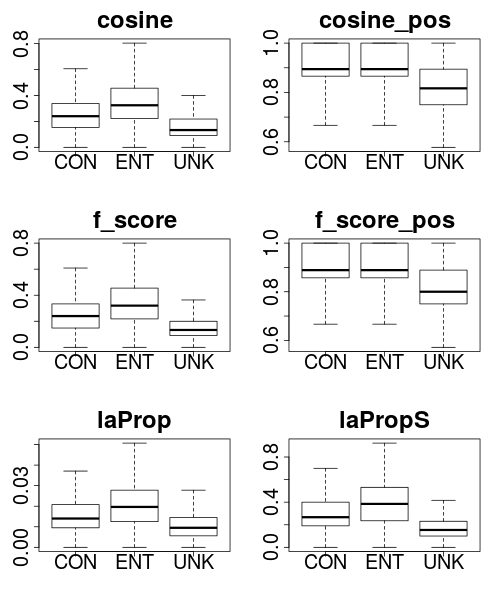}
\caption{Distributions of the similarity metrics by tweet pair class for the event {\em ssiege} in the {\em Threads} ({\bf left}) and the {\em iPosts} dataset (\bf right).}
\label{fig:stat2}
\end{center}
\end{figure}

Figures \ref{fig:stat1} and \ref{fig:stat2} show the distribution of the features introduced above each for a selected event in both datasets. Each figure half
represents a dataset;  each subplot shows the distribution of a feature in dependence of the three RTE classes for the selected event in that dataset.

The plots indicate a general trend over all events and datasets:  the similarity features reach highest values for the ENT class, followed by CON and UNK. Kruskal-Wallis tests applied separately for all combinations of features, events and datasets confirmed these trends, revealing significant differences for all boxplot triplets ($p<0.001$ after correction for type 1 errors in this high amount of comparisons using the false discovery rate method of \cite{Benjamini2001}). 
Dunnett post hoc tests however clarified that for 16 out of 72 comparisons (all POS similarity measures) only UNK but not ENT and CON differ significantly ($\alpha=0.05$). Both datasets contain the same amount of non-significant cases. Nevertheless, these trends are encouraging to test whether an RTE task can be addressed by string and POS-level similarity features alone, without 
syntactic or semantic level tweet comparison.

\section{RTE classification experiments for Contradiction and Disagreeing Reply detection}
\label{sec:method}

In order to predict the RTE classes based on the features introduced above, we trained two classifiers: Nearest (shrunken) centroids (NC) \cite{Tibshirani2003} and Random forest (RF) \cite{Breiman2001,Liaw2002}, using the R wrapper package {\em Caret} \cite{Kuhn2016} with the methods {\em pam} and {\em rf}, respectively. 
To derive the same number of instances for all classes, we applied separately for both datasets resampling without replacement, so that the total data amounts about 4,550 feature vectors equally distributed over the three classes, the majority of 4,130 belonging to the iPosts data set. 
Further, we centered and scaled the feature matrix.
Within the Caret framework we optimized the tunable parameters of both classifiers by maximizing the F1 score. This way the NC shrinkage delta was set to 0, which means that the class reference centroids are not modified. 
For RF the number of variables randomly sampled as candidates at each split was set to 2.
The remaining parameters were kept default.

The classifiers were tested on both datasets in a 4-fold event-based held-out setting, training on three events and testing on the remaining one (4-fold cross-validation, CV),  quantifying how performance generalizes to new events with unseen claims and unseen targets.
The CV scores are summarized in Tables \ref{tab:res1} and \ref{tab:res2}. It turns out generally that classifying CON is more difficult than classifying ENT or UNK.
We observe  a dependency of the classifier performances on the two contradiction scenarios: for detecting CON, RF achieved higher classification values on  Threads, 
whereas NC performed better on iPosts. 
General performance across all three classes was better in independent posts  
than in conversational threads.

Definitions of contradiction,  the genre of texts and the features used are  dependent on end applications, 
making performance comparison nontrivial \cite{pheme_deliv_422}. On a different subset of the Threads data in terms of events, size of evidence, 4 stance classes and no resampling,
\cite{Lukasik} report .40 overall F-score using Gaussian processes, cosine similarity on text vector representation and temporal metadata. 
 Our previous experiments were done  using the Excitement Open Platform  incorporating syntactico-semantic processing and 4-fold CV. For the non-resampled Threads data we reported .11 F1 on CON  via training on iPosts \cite{pheme_deliv_422}.  On the non-resampled iPosts data we obtained .51 overall  F1 score 
\cite{Lendvaietal_pheme_rte}, F1  on CON being .25 \cite{pheme_deliv_422}.

\vfill

\begin{table}[!ht]
\begin{center}
\begin{tabular}{|r||c|c|c|}
\hline  
    & CON  & ENT  &  UNK \\
\hline
\hline
F1 (RF/{\bf NC}) & 0.33/{\bf 0.35} & 0.55/0.59 & 0.51/0.57 \\
precision & 0.35/0.40 & 0.54/0.61 & 0.54/0.57 \\
recall  & 0.32/0.34 & 0.58/0.59 & 0.56/0.67 \\
\hline
accuracy & \multicolumn{3}{|c|}{0.47/0.51} \\
wgt F1 & \multicolumn{3}{|c|}{0.48/{\bf 0.51}} \\
wgt prec. & \multicolumn{3}{|c|}{0.51/0.55} \\
wgt rec. & \multicolumn{3}{|c|}{0.47/0.51} \\
\hline
\end{tabular}
\end{center}
\caption{{\em iPosts} dataset. Mean and weighted (wgt) mean results on held-out data after event held-out cross validation for the Random Forest (RF) and Nearest Centroid (NC) classifiers.}
\label{tab:res1}
\end{table}

\begin{table}[!ht]
\begin{center}
\begin{tabular}{|r||c|c|c|}
\hline  
    & CON  & ENT  &  UNK \\
\hline
\hline
F1 ({\bf RF}/NC) & {\bf 0.37}/0.11 & 0.45/0.50 & 0.40/0.36 \\
precision & 0.42/0.07 & 0.52/0.56 & 0.34/0.31 \\
recall  & 0.35/0.20 & 0.41/0.47 & 0.50/0.61 \\
\hline
accuracy & \multicolumn{3}{|c|}{0.42/0.39} \\
wgt F1 & \multicolumn{3}{|c|}{{\bf 0.43}/0.32} \\
wgt prec. & \multicolumn{3}{|c|}{0.47/0.33} \\
wgt rec. & \multicolumn{3}{|c|}{0.42/0.39} \\
\hline
\end{tabular}
\end{center}
\caption{{\em  Threads} dataset. Mean and weighted (wgt) mean results on held-out data after event held-out cross validation for the Random forest and Nearest Centroid classifiers (RF/NC).}
\label{tab:res2}
\end{table}

We proposed to model two types of contradictions: in the first both tweets encode the claim target (iPosts), in the second typically only one of them (Threads). The Nearest Centroid algorithm  performs   poorly  on the CON class in Threads where  textual overlap  is typically small  especially for the CON and UNK  classes,  in part due to the absence of the claim target in replies.  However, the Random Forest algorithm's performance is not affected by this factor.
The advantage of RF on the Threads data can be explained by its property of training several weak classifiers on parts of the feature vectors only. By this boosting strategy a usually undesirable combination of relatively long feature vectors but few training observations can be tackled, holding for the Threads data that due to its extreme skewedness (cf. Table \ref{tab:collections}) shrunk down to only 420 datapoints after our class balancing technique of resampling without replacement. Results indicate the benefit of RF classifiers in such sparse data cases.

The good performance of NC on the much larger amount of data in iPosts
is in line with the corpus statistics
reported in section \ref{sec:corpstat}, implying a reasonably small amount of class overlap. The classes are thus relatively well represented by their centroids, which is exploited by the NC classifier.
However, as illustrated in Figures \ref{fig:stat1} and \ref{fig:stat2}, the majority of feature distributions 
are generally better separated for ENT and UNK, while
CON in its mid position shows more overlap to both other classes and is thus overall a less distinct category.

\section{Conclusions and Future Work}
\label{sec:conc}

The detection of  contradiction and disagreement in microposts  supplies important cues to factuality and veracity assessment, and is a central task in computational journalism. 
We  developed classifiers in a uniform, general inference framework that differentiates two tasks based on  contextual proximity of the two posts to be assessed, and if the claim target may or may not be omitted in their content. We utilized simple text similarity metrics that proved to be a good basis for  contradiction classification.

Text similarity was measured in terms of vocabulary and token sequence overlap. To derive the latter, local alignment turned out to be a valuable tool: as opposed to standard global alignment \cite{Wagner1974}, it can account for crossing dependencies and thus for varying sequential order of information structure in entailing text pairs, e.g. in "the cat chased the mouse" and "the mouse was chased by the cat", which are differently structured into topic and comment \cite{Halliday1967}. 
We expect 
contradictory content 
to exhibit similar trends in variation with respect to 
content unit order -- especially in the Threads scenario, where entailment inferred from a reply can become the topic of a subsequent replying tweet. Since local alignment can resolve such word order differences, it is able to preserve  text similarity of entailing tweet pairs, which is reflected in the relative {\em laProp} boxplot heights in Figures \ref{fig:stat1} and \ref{fig:stat2}.

We have run leave-one-event-out evaluation separately on the independent posts data and on the conversational threads data, which allowed us to compare performances on  collections originating from the same genre and  platform, but on content where claim targets in the test data are different from the targets in the training data.  Our obtained generalization performance over unseen events turns out to be in line with previous reports.
Via downsampling, we achieved a balanced performance on both tasks across the three RTE classes; however, in line with previous work, even in this setup the overall performance on contradiction is the lowest, whereas detecting the lack of contradiction can be achieved with much better performance in both contradiction scenarios.

Possible extensions to our approach include  incorporating more informed text similarity metrics \cite{ukp_similarity}, formatting phenomena  \cite{Tolosi}, and distributed contextual representations \cite{doc_embeddings}, the 
utilization of knowledge-intensive resources \cite{pado_eop}, representation of  alignment on various content levels \cite{noh_multilevel}, and formalization of  contradiction scenarios in terms of additional  layers of perspective  \cite{perspectives_annotation}.

\section{Acknowledgments}

P. Lendvai was supported  by  the PHEME FP7 project (grant nr. 611233), U. D. Reichel by an Alexander von Humboldt Society grant. We thank anonymous reviewers for their input.

\newpage

\bibliographystyle{acl}
\bibliography{piro_newbib}

\begin{thebibliography}{}

\bibitem[\protect\citename{Agirre \bgroup et al.\egroup
  }2016]{semeval_interpretable}
Eneko Agirre, Aitor Gonzalez-Agirre, Inigo Lopez-Gazpio, Montse Maritxalar,
  German Rigau, and Larraitz Uria.
\newblock 2016.
\newblock {Semeval-2016 task 2: Interpretable semantic textual similarity}.
\newblock {\em Proceedings of SemEval}, pages 512--524.

\bibitem[\protect\citename{Augenstein \bgroup et al.\egroup
  }2016]{Augenstein2016SemEval}
Isabelle Augenstein, Andreas Vlachos, and Kalina Bontcheva.
\newblock 2016.
\newblock {USFD: Any-Target Stance Detection on Twitter with Autoencoders}.
\newblock In Saif~M. Mohammad, Svetlana Kiritchenko, Parinaz Sobhani, Xiaodan
  Zhu, and Colin Cherry, editors, {\em Proceedings of the International
  Workshop on Semantic Evaluation}, SemEval '16, San Diego, California.

\bibitem[\protect\citename{B{\"a}r \bgroup et al.\egroup }2012]{ukp_similarity}
Daniel B{\"a}r, Chris Biemann, Iryna Gurevych, and Torsten Zesch.
\newblock 2012.
\newblock {UKP}: Computing semantic textual similarity by combining multiple
  content similarity measures.
\newblock In {\em Proceedings of the First Joint Conference on Lexical and
  Computational Semantics-Volume 1: Proceedings of the main conference and the
  shared task, and Volume 2: Proceedings of the Sixth International Workshop on
  Semantic Evaluation}, pages 435--440. Association for Computational
  Linguistics.

\bibitem[\protect\citename{Benjamini and Yekutieli}2001]{Benjamini2001}
Yoav Benjamini and Daniel Yekutieli.
\newblock 2001.
\newblock The control of the false discovery rate in multiple testing under
  dependency.
\newblock {\em Annals of Statistics}, 29:1165--1188.

\bibitem[\protect\citename{Bowman \bgroup et al.\egroup }2015]{Bowman}
Samuel~R. Bowman, Gabor Angeli, Christopher Potts, and Christopher~D. Manning.
\newblock 2015.
\newblock A large annotated corpus for learning natural language inference.
\newblock In {\em Proceedings of the 2015 Conference on Empirical Methods in
  Natural Language Processing}, pages 632--642, Lisbon, Portugal, September.
  Association for Computational Linguistics.

\bibitem[\protect\citename{Breiman}2001]{Breiman2001}
Leo Breiman.
\newblock 2001.
\newblock Random forests.
\newblock {\em Machine Learning}, 45(1):5--32.

\bibitem[\protect\citename{Dagan \bgroup et al.\egroup }2006]{dagan_pascal}
Ido Dagan, Oren Glickman, and Bernardo Magnini.
\newblock 2006.
\newblock {The PASCAL recognising textual entailment challenge}.
\newblock In {\em Machine learning challenges. evaluating predictive
  uncertainty, visual object classification, and recognising tectual
  entailment}, pages 177--190. Springer.

\bibitem[\protect\citename{De~Marneffe \bgroup et al.\egroup }2008]{deMarneffe}
Marie-Catherine De~Marneffe, Anna~N Rafferty, and Christopher~D Manning.
\newblock 2008.
\newblock Finding contradictions in text.
\newblock In {\em Proc. of ACL}, volume~8, pages 1039--1047.

\bibitem[\protect\citename{de Marneffe \bgroup et al.\egroup
  }2012]{deMarneffe_didithappen}
Marie-Catherine de~Marneffe, Christopher~D Manning, and Christopher Potts.
\newblock 2012.
\newblock Did it happen? the pragmatic complexity of veridicality assessment.
\newblock {\em Computational Linguistics}, 38(2):301--333.

\bibitem[\protect\citename{Ferreira and Vlachos}2016]{Emergent}
William Ferreira and Andreas Vlachos.
\newblock 2016.
\newblock {Emergent: a novel data-set for stance classification}.
\newblock In {\em Proceedings of NAACL}.

\bibitem[\protect\citename{Halliday}1967]{Halliday1967}
Michael Alexander~Kirkwood Halliday.
\newblock 1967.
\newblock Notes on transitivity and theme in {E}nglish, part {II}.
\newblock {\em Journal of Linguistics}, 3(2):199--244.

\bibitem[\protect\citename{Kuhn}2016]{Kuhn2016}
Max Kuhn, 2016.
\newblock {\em caret: Classification and Regression Training}.
\newblock R package version 6.0-71.

\bibitem[\protect\citename{Le and Mikolov}2014]{doc_embeddings}
Quoc~V Le and Tomas Mikolov.
\newblock 2014.
\newblock Distributed representations of sentences and documents.
\newblock In {\em ICML}, volume~14, pages 1188--1196.

\bibitem[\protect\citename{Lendvai \bgroup et al.\egroup
  }2016a]{Lendvaietal_pheme_rte}
Piroska Lendvai, Isabelle Augenstein, Kalina Bontcheva, and Thierry Declerck.
\newblock 2016a.
\newblock Monolingual social media datasets for detecting contradiction and
  entailment.
\newblock In {\em Proc. of LREC-2016}.

\bibitem[\protect\citename{Lendvai \bgroup et al.\egroup
  }2016b]{pheme_deliv_422}
Piroska Lendvai, Isabelle Augenstein, Dominic Rout, Kalina Bontcheva, and
  Thierry Declerck.
\newblock 2016b.
\newblock {Algorithms for Detecting Disputed Information. Deliverable D4.2.2
  for FP7-ICT Collaborative Project ICT-2013-611233 PHEME}.
\newblock \url{https://www.pheme.eu/wp-content/uploads/2016/06/D422_final.pdf}.

\bibitem[\protect\citename{Liaw and Wiener}2002]{Liaw2002}
Andy Liaw and Matthew Wiener.
\newblock 2002.
\newblock Classification and regression by random{F}orest.
\newblock {\em R News}, 2(3):18--22.

\bibitem[\protect\citename{Lukasik \bgroup et al.\egroup }2016]{Lukasik}
Michal Lukasik, P.K. Srijith, Duy Vu, Kalina Bontcheva, Arkaitz Zubiaga, and
  Trevor Cohn.
\newblock 2016.
\newblock {Hawkes Processes for Continuous Time Sequence Classification: An
  Application to Rumour Stance Classification in Twitter}.
\newblock In {\em Proceedings of ACL-16}.

\bibitem[\protect\citename{Madhumita}2016]{madhumita}
Madhumita.
\newblock 2016.
\newblock Recognizing textual entailment.
\newblock Master's thesis, Saarland University, Saarbr\"ucken, Germany.

\bibitem[\protect\citename{Marcus \bgroup et al.\egroup }1999]{Penn1999}
Mitchell~P. Marcus, Ann Taylor, Robert MacIntyre, Ann Bies, Constance Cooper,
  Mark Ferguson, and Alison Littman.
\newblock 1999.
\newblock {The Penn Treebank Project}.
\newblock \url{http://www.cis.upenn.edu/~treebank/home.html}.
\newblock visited on Sep 29th 2016.

\bibitem[\protect\citename{Mendoza \bgroup et al.\egroup }2010]{Mendoza2010}
Marcelo Mendoza, Barbara Poblete, and Carlos Castillo.
\newblock 2010.
\newblock {Twitter Under Crisis: Can We Trust What We RT?}
\newblock In {\em Proceedings of the First Workshop on Social Media Analytics
  (SOMA'2010)}, pages 71--79, New York, NY, USA. ACM.

\bibitem[\protect\citename{Mohammad \bgroup et al.\egroup
  }2016]{StanceSemEval2016}
Saif~M. Mohammad, Svetlana Kiritchenko, Parinaz Sobhani, Xiaodan Zhu, and Colin
  Cherry.
\newblock 2016.
\newblock {SemEval-2016 Task 6}: Detecting stance in tweets.
\newblock In {\em Proceedings of the International Workshop on Semantic
  Evaluation}, SemEval '16, San Diego, California.

\bibitem[\protect\citename{Morante and Blanco}2012]{morante_scope}
Roser Morante and Eduardo Blanco.
\newblock 2012.
\newblock {*SEM 2012 shared task: Resolving the scope and focus of negation}.
\newblock In {\em Proceedings of the First Joint Conference on Lexical and
  Computational Semantics}.

\bibitem[\protect\citename{Morante and Sporleder}2012]{exprom}
Roser Morante and Caroline Sporleder, editors.
\newblock 2012.
\newblock {\em {ExProm '12: Proceedings of the ACL-2012 Workshop on
  Extra-Propositional Aspects of Meaning in Computational Linguistics}}.
  Association for Computational Linguistics.

\bibitem[\protect\citename{Noh \bgroup et al.\egroup }2015]{noh_multilevel}
Tae-Gil Noh, Sebastian Pad{\'o}, Vered Shwartz, Ido Dagan, Vivi Nastase,
  Kathrin Eichler, Lili Kotlerman, and Meni Adler.
\newblock 2015.
\newblock Multi-level alignments as an extensible representation basis for
  textual entailment algorithms.
\newblock {\em Lexical and Computational Semantics (* SEM 2015)}, page 193.

\bibitem[\protect\citename{Pad\'o \bgroup et al.\egroup }2015]{pado_eop}
Sebastian Pad\'o, Tae-Gil Noh, Asher Stern, Rui Wang, and Roberto Zanoli.
\newblock 2015.
\newblock {Design and Realization of a Modular Architecture for Textual
  Entailment}.
\newblock {\em Natural Language Engineering}, 21(02):167--200.

\bibitem[\protect\citename{Porter}1980]{Porter1980}
Martin~F. Porter.
\newblock 1980.
\newblock An algorithm for suffix stripping.
\newblock {\em Program}, 14(3):130--137.

\bibitem[\protect\citename{Qazvinian \bgroup et al.\egroup
  }2011]{Qazvinian2011}
Vahed Qazvinian, Emily Rosengren, Dragomir~R. Radev, and Qiaozhu Mei.
\newblock 2011.
\newblock Rumor has it: Identifying misinformation in microblogs.
\newblock In {\em Proceedings of the Conference on Empirical Methods in Natural
  Language Processing}, EMNLP '11, pages 1589--1599.

\bibitem[\protect\citename{Reichel}2012]{ReichelIS2012}
Uwe~D. Reichel.
\newblock 2012.
\newblock Perm{A} and {B}alloon: {T}ools for string alignment and text
  processing.
\newblock In {\em Proc. Interspeech}, page paper no. 346, Portland, Oregon,
  USA.

\bibitem[\protect\citename{Smith and Waterman}1981]{SmithJMB1981}
Temple~F. Smith and Michael~S. Waterman.
\newblock 1981.
\newblock Identification of common molecular subsequences.
\newblock {\em Journal of Molecular Biology}, 147:195--197.

\bibitem[\protect\citename{Tibshirani \bgroup et al.\egroup
  }2003]{Tibshirani2003}
Robert Tibshirani, Trevor Hastie, Balasubramanian Narasimhan, and Gilbert Chu.
\newblock 2003.
\newblock Class prediction by nearest shrunken centroids,with applications to
  {DNA} microarrays.
\newblock {\em Statistical Science}, 18(1):104--117.

\bibitem[\protect\citename{Tolosi \bgroup et al.\egroup }2016]{Tolosi}
Laura Tolosi, Andrey Tagarev, and Georgi Georgiev.
\newblock 2016.
\newblock An analysis of event-agnostic features for rumour classification in
  twitter.
\newblock In {\em Proc. of Social Media in the Newsroom Workshop}.

\bibitem[\protect\citename{van Son \bgroup et al.\egroup
  }2016]{perspectives_annotation}
Chantal van Son, Tommaso Caselli, Antske Fokkens, Isa Maks, Roser Morante, Lora
  Aroyo, and Piek Vossen.
\newblock 2016.
\newblock {GRaSP: A Multilayered Annotation Scheme for Perspectives}.
\newblock In {\em Proceedings of the 10th Edition of the Language Resources and
  Evaluation Conference (LREC)}.

\bibitem[\protect\citename{Wagner and Fischer}1974]{Wagner1974}
Robert~A. Wagner and Michael~J. Fischer.
\newblock 1974.
\newblock {T}he string-to-string correction problem.
\newblock {\em Journal of the Association for Computing Machinery},
  21(1):168--173.

\bibitem[\protect\citename{Wang and Neumann}2007]{wang2007recognizing}
Rui Wang and G{\"u}nter Neumann.
\newblock 2007.
\newblock Recognizing textual entailment using a subsequence kernel method.
\newblock In {\em AAAI}, volume~7, pages 937--945.

\bibitem[\protect\citename{Xu \bgroup et al.\egroup }2015]{2015semeval}
Wei Xu, Chris Callison-Burch, and William~B Dolan.
\newblock 2015.
\newblock {SemEval-2015 Task 1: Paraphrase and semantic similarity in Twitter
  (PIT)}.
\newblock {\em Proceedings of SemEval}.

\bibitem[\protect\citename{Zubiaga \bgroup et al.\egroup }2015]{Zubiaga}
Arkaitz Zubiaga, Maria Liakata, Rob Procter, Kalina Bontcheva, and Peter
  Tolmie.
\newblock 2015.
\newblock {Towards Detecting Rumours in Social Media}.
\newblock {\em CoRR}, abs/1504.04712.

\end{thebibliography}

\end{document}